%% file: main.tex
\documentclass{article}
\usepackage[a4paper, margin=2cm]{geometry}

\usepackage{natbib}

\usepackage{graphicx}
\usepackage{subfigure}

\usepackage[utf8]{inputenc}
\usepackage[T1]{fontenc}
\usepackage[svgnames]{xcolor}
\usepackage[colorlinks=true,linkcolor=Maroon, urlcolor=Maroon,citecolor=RoyalBlue,backref=page]{hyperref}
\usepackage{url}
\usepackage{booktabs}
\usepackage{amsfonts}      
\usepackage{nicefrac}      
\usepackage{microtype}     

\usepackage{amsmath}
\usepackage{amssymb}
\usepackage{mathtools}
\usepackage{amsthm}
\usepackage{tikz}
\usetikzlibrary{shapes,external}

\usepackage{./artem}
\usepackage{./additional}

\usepackage{amssymb}
\usepackage{enumitem}

\usepackage[frozencache,cachedir=.]{minted}
\usepackage{siunitx}
\usepackage{dirtytalk}
\usepackage{wasysym}

\usepackage{algorithm}
\usepackage[noend]{algorithmic}

\newcommand\algorithmicprocedure{\textbf{procedure}}
\newcommand{\algorithmicendprocedure}{\algorithmicend\ \algorithmicprocedure}
\makeatletter
\newcommand\PROCEDURE[3][default]{%
  \ALC@it
  \algorithmicprocedure\ \textsc{#2}(#3)%
  \ALC@com{#1}%
  \begin{ALC@prc}%
}
\newcommand\ENDPROCEDURE{%
  \end{ALC@prc}%
  \ifthenelse{\boolean{ALC@noend}}{}{%
    \ALC@it\algorithmicendprocedure
  }%
}
\newenvironment{ALC@prc}{\begin{ALC@g}}{\end{ALC@g}}
\makeatother

\newlength\myindent
\makeatletter
\AtBeginEnvironment{minted}{\dontdofcolorbox}
\def\dontdofcolorbox{\renewcommand\fcolorbox[4][]{##4}}
\makeatother

\tikzset{every picture/.style={line width=0.75pt}}

\usepackage{authblk}

\makeatletter
\patchcmd{\@maketitle}{\LARGE \@title}{\fontsize{16}{19.2}\selectfont\@title}{}{}
\makeatother

\title{\textbf{Memory Safe Computations with XLA Compiler}}

\author[1,2]{Artem Artemev}
\author[1]{Tilman Roeder}
\author[1]{Mark van der Wilk}
\affil[ ]{{\{a.artemev20, tilman.roeder17, m.vdwilk\}@imperial.ac.uk}}
\affil[1]{Imperial College London}
\affil[2]{Secondmind}
\date{}

\begin{document}

\maketitle

\begin{abstract}
Software packages like TensorFlow and PyTorch are designed to support linear algebra operations, and their speed and usability determine their success. However, by prioritising speed, they often neglect memory requirements. As a consequence, the implementations of memory-intensive algorithms that are convenient in terms of software design can often not be run for large problems due to memory overflows. Memory-efficient solutions require complex programming approaches with significant logic outside the computational framework. This impairs the adoption and use of such algorithms. To address this, we developed an XLA compiler extension\footnote{The code is available at \url{https://github.com/awav/tensorflow}} that adjusts the computational data-flow representation of an algorithm according to a user-specified memory limit. We show that k-nearest neighbour and sparse Gaussian process regression methods can be run at a much larger scale on a single device, where standard implementations would have failed. Our approach leads to better use of hardware resources. We believe that further focus on removing memory constraints at a compiler level will widen the range of machine learning methods that can be developed in the future.
\end{abstract}

\section{Introduction}
\label{sec:intro}
Progress in science is inextricably linked with advances in scientific computing, in terms of both software and hardware. This is particularly noticeable in machine learning through the huge impact of numerical software packages supporting automatic differentiation \citep{baydin2018automatic}. Packages such as TensorFlow \citep{abadi2016tensorflow}, PyTorch \citep{paszke2019pytorch}, or JAX \citep{jax2018github} greatly accelerated \textbf{1)} the implementation of gradient-based optimisation procedures by eliminating error-prone manual differentiation, and \textbf{2)} the execution of code by leveraging modern and heterogeneous hardware (e.g.~GPU, TPU or IPU). A large portion of this impact is attributable to the accessible and user-friendly form that these features were delivered in. This contributed to the growth in the machine learning community, in terms of methodological researchers, as well as the wider scientific audience and practitioners.

The aforementioned software frameworks work by chaining together efficient implementations of mathematical operations (known as \textit{kernels}). By providing implementations that are tailored to various types of hardware, a speed-optimised implementation can be obtained. While speed is certainly important to pursue, many algorithms face a different challenge: hardware memory constraints. Often, these have a larger impact, as memory constraint violations can lead to the execution terminating before an answer is obtained. This make-or-break property is particularly noticeable on GPUs, where allocating more memory than is physically available leads to an immediate termination of execution, and larger amounts of physical memory comes at a significant cost. 

Now that numerical computation frameworks are widely used, they strongly influence what machine learning algorithms are adopted. This happens through hard limitations, as well as usability considerations through what is easily implementable. Currently, the emphasis on optimising runtime causes many algorithms to be severely memory limited, or too cumbersome to implement. This is particularly noticeable in methods that rely heavily on matrix and linear algebra computations, e.g.~kernel methods \citep[e.g.][]{titsias2009variational} or nearest neighbour methods for geometric deep learning \citep{bronstein2017geometric}.

In this work, we aim to remove these limitations, by developing a tool that optimises code to be more memory efficient, with a particular focus on linear algebra operations. This optimisation is transparent to the user, and therefore allows many algorithms to be run at scales that were previously impossible, while leaving implementations as simple as before. This allows a wider range of algorithms to take advantage of ``the bitter lesson''---\textit{\say{General methods that leverage computation are ultimately the most effective}} \citep{sutton2019bitter}---while making them more accessible to the wider community, as computational frameworks have sought to do all along.

Our method is implemented as an extension to the XLA compiler \citep{leary2017xla}, which we chose due to its wide use and support for optimising computations specified in TensorFlow and JAX. We demonstrate the benefits of our method by scaling algorithms where simple implementations do not scale due to memory bottlenecks, such as k-nearest-neighbours, and sparse Gaussian process regression \citep{titsias2009variational}. With our extensions, these methods scale to far larger problems, \textit{without changing a single line} of their implementation in Python. Our Gaussian process experiment shows that simply scaling up a 13 year old method can outperform much more recent methods, indicating that older methods may be undervalued in recent literature.

\section{Motivation: Memory-Constrained Machine Learning}
Since memory overflows cause the execution of code to be immediately halted without producing any result, memory constraints form the key obstacle for scaling many machine learning algorithms. In addition, memory is a scarce resource that comes at a considerable cost, particularly in GPUs. This causes memory to be a key limiting factor for researchers and practitioners using machine learning tools at scale.
This is particularly noticeable in algorithms where minibatching is undesirable and that rely on pairwise distances, like k-Nearest Neighbours (kNN) or Gaussian processes \citep{rasmussen2006gaussian} (which we particularly focus on in this work). Even in modern deep learning, memory constraints cause problems by limiting batch sizes, layer widths, or sizes of attention mechanisms \citep{vaswani2017transformer}. 
In all of these examples, matrix and linear algebra operations cause the bottleneck. For kNN, kernel/GP methods, and transformers the root of the problem is a pairwise matrix needs to be computed between inputs, giving a quadratic memory cost.

Often, more memory efficient implementations \textit{can} be programmed at the cost of increased software complexity.
This ranges from minor annoyances, for example accumulating minibatch gradients in an outer loop for large-batch training, to complex engineering efforts that have been published as scientific contributions in their own right, for example in scaling Gaussian processes to $>10^5$ datapoints \citep{gal2014distributed,wang2019exact,meanti2020falcon}.

Our goal is to provide a tool that finds memory-efficient ways to execute algorithms, without the need for increasing software complexity. This will allow scientists and practitioners to access the benefits of scale in existing methods more easily, and without incurring the cost of expensive large-memory hardware. For the main demonstration of our approach, we will automatically obtain a memory-efficient implementation of sparse Gaussian process regression \citep{titsias2009variational}, which was previously implemented with considerable difficulty \citep{gal2014distributed}. The increase in scale makes the method competitive in comparisons where it was previously dismissed as not scalable enough \citep{wang2019exact}, showing the value of reducing the barriers to scaling.

\section{Related Work}
\label{sec:background}
A popular approach to address memory issues is distributing computation across multiple resources like a group of GPUs or a computer cluster with network protocol connectivity between machines \citep{buyya1999high,dean2008mapreduce}. More specifically, sharding allows large tensors to be split up and distributed across multiple devices, which increases the total amount of memory available for an algorithm, but comes at the cost of requiring more hardware resources. Most computational frameworks\footnote{Published under permissive open-source licenses, like Apache or BSD.}
\citep{abadi2016tensorflow,shazeer2018mesh,jax2018github,paszke2019pytorch} support sharding, although some manual specification of how to distribute computation is required.
This complicates an implementation and requires the user to have a wide engineering skill set.
An automatic sharding tool such as Tofu \citep{wang2019tofu} eases user implementation experience. Although, Tofu does require the user to specify their computations through a custom interface, which requires modifying code.
While sharding approach does allow scaling of certain implementations, it remains wasteful for algorithms that \textit{can} be implemented in a more memory-efficient way, but where it is simply cumbersome to do so.

Compilers have been introduced to allow humans to express programs in an elegant way, while generating programs that actually run well on specific hardware \cite{aho2006compilers}. Our goal of obtaining memory-efficient implementations, while keeping code convenient for humans, is therefore suited to be addressed by adding memory optimisations to a compiler. Compilers are already being used to optimise computational graphs, notably in JAX, TensorFlow and PyTorch by XLA \citep{leary2017xla}, TVM \citep{chen2018tvm}, Glow \citep{rotem2018glow} for PyTorch only. TVM performs similar optimisations to XLA, but unlike XLA, it is not seamlessly integrated into popular frameworks and requires additional user effort.

The optimisations in XLA mainly focus on increasing code speed, for example through \textit{common sub-expression elimination} (CSE), \textit{dead code elimination} (DCE), operations \textit{fusion}, and other more specific modifications. The main advantage of XLA is that it optimises computations in a way that is completely transparent to the user who specifies the computational graph. Although XLA and TVM implement low-level memory optimisations, they do not adapt code handling large tensors to satisfy memory constraints.
For the matrix and linear algebra tasks that we consider, KeOps \citep{feydy2020fast,charlier2021keops} currently provides the most efficient memory management. To achieve any benefits, a user must specify a series of computations using KeOps classes, which form a layer above the PyTorch framework. KeOps works similarly to a compiler, by first building a symbolic representation of the computation, which allows the computation to be broken into memory-efficient sections, that are then run with custom CUDA kernels.

In terms of prior work, KeOps is closest in aim and achievement to ours. We aim to address three of its limitations. Firstly, KeOps requires users to reimplement their algorithms using KeOps classes. While the programming interface is elegant, needing to mix KeOps and other computational frameworks does add complexity. Secondly, for KeOps to be able to optimise an operation, it has to be reimplemented within KeOps, which significantly duplicates effort. Finally, because of the former drawback, KeOps does not inherit the support for a wide range of hardware from e.g.~JAX/TensorFlow.

\section{Memory Efficient Matrix and Linear Algebra Operations in XLA}
\label{sec:exla-optimisations}
Compilers are a highly promising way for improving runtime properties of code, without requiring user intervention and while leaving code elegant. The specific matrix and linear algebra optimisations that we consider have not yet been implemented in any of the frameworks discussed above. However, they \textit{could} be implemented in any of TVM, KeOps, or XLA. We choose to extend XLA over TVM, because of XLA's better integration with common computational frameworks. In addition, we choose to extend XLA over KeOps, because it \textbf{1)} does not require algorithms to be rewritten in a separate framework, \textbf{2)} can optimise computational graphs in their entirety, rather than just what is implemented in the separate framework, and \textbf{3)} can take advantage of the full capabilities that already exist in JAX/TensorFlow.

We introduce several optimisation strategies (known as \textit{optimisation passes} in the XLA codebase) into the XLA optimisation pipeline. We aim to constrain the program's memory footprint with minimal sacrifices in the execution speed. The optimisation passes examine the entire computational data-flow graph (High Level Optimiser Internal Representation, or HLO IR), search for weak spots, and try to eliminate them. Abstractions at a similar level to HLO IR have been shown to be convenient for optimising linear algebra operations \citep{barthels2021linnea}.
We add match-and-replace operations, e.g.~to introduce a more efficient distance computation, reshuffling operations for expressions that are invariant to evaluation order, and splitting with large tensors to reduce memory usage.

\subsection{Match and replace}
\label{ssec:match-replace}
The \textit{match and replace} optimisation pass searches for expressions in a data-flow graph for which we know in advance that an equivalent and more efficient version exists. For example, we search for expressions that compute Euclidean distance in naive form between vectors of length $n$ and $m$ with a dimension $d$. The naive Euclidean distance computation uses broadcasting over the dimension $d$ and creates a temporary tensor with entries $(x_{nd} - y_{md})^2$ of size $n \times m \times d$. This can be replaced with $\sum_d x_{nd}^2 + y_{md}^2 - 2x_{nd}y_{md}$, where the largest tensor has size $n \times m$.

Replacing sub-parts of the graph is a standard procedure in compilers like XLA, although many linear algebra tricks have been missing. While the Euclidean distance is the only match-and-replace optimisation we implement, other operations can easily be added, for example, efficiently adding diagonals to matrices without allocating a dense square tensor where only the diagonal is non-zero.

\subsection{Reordering}
\label{ssec:reordering}
\begin{listing}[t]
\begin{minted}{python}
@jax.jit
def matrix_matrix_vector_mul(A, B, v):
    C = A @ B @ v
    return C
\end{minted}
\caption{Chain multiplication example $C = ABv$ for $A, B \in \R^{n \times n}$, and $v \in \R^n$.}
\label{code:chain-jax}
\end{listing}
A computational data-flow graph is an ordered sequence of operations, with the order of operations influencing the memory usage. In some cases, reordering sub-parts of the data-flow graph can lead to reductions in the memory footprint.
The classical example of reordering is the optimisation of matrix chain multiplications. For example, consider the matrix expression $\textup{C} = \textup{A}\textup{B}\bfv$ for matrices $\textup{A}, \textup{B} \in \R^{n \times n}$, and $\bfv \in \R^n$. In the \cref{code:chain-jax}, the order of operations determines that the matrix multiplications are performed from left to right, i.e.~$\textup{C} = {\color{red}(\textup{A}\textup{B})}\bfv$, which gives the most inefficient execution order with runtime complexity $O(n^3)$ and memory complexity $O(n^2)$. Changing the order to $\textup{C} = \textup{A} {\color{red}(\textup{B} \bfv)}$ improves time complexity to $O(n^2)$ and practical memory complexity because the intermediate multiplication result of $\color{red}\textup{B}\bfv$ is a vector not a matrix as in the case of $\color{red}\textup{A}\textup{B}$ multiplication.

The optimisation of matrix chain multiplication is possible due to the associativity of matrix multiplication, such that the result of the matrix multiplication chain does not depend on where parentheses are placed.
There are many efficient and sophisticated algorithms for addressing this task \citep{chin1978n,czumaj1996very,barthels2018generalized,schwartz2019revisit}.
We implement a simplified procedure for reordering matrix vector chain multiplications that detects inefficient matrix multiplication chains, which are guaranteed to reduce in size at the end of the chain.

\subsection{Data-flow graph splitting}
\label{ssec:splitting}

\begin{algorithm}[h!]
\caption{High-level description of the depth-first search visitor-handler that splits the data-flow graph up to the reduction \texttt{dot} operation. Symbol $\rightsquigarrow$ denotes a directed computational path in the data-flow graph. Steps \numrange[range-phrase = --]{9}{12} are done recursively traversing back visited operations in the data-flow graph.}
\begin{algorithmic}[1]
\PROCEDURE{HandleDot}{\texttt{dot}: \texttt{HloInstruction}}
\IF{$\texttt{output\_size}(\texttt{dot}) \ge \texttt{tensor\_size\_threshold}$, i.e. \texttt{dot} is splittable}
    \STATE Exit and continue traversing succeeding operations in the data-flow graph and search for size-reducing operation for the output tensor of \texttt{dot}.
\ENDIF
\IF{\texttt{dot.rhs} \textit{is not} splittable and \texttt{dot.lhs} \textit{is not} splittable}
    \STATE Exit and continue traversing the data-flow graph.
\ENDIF
\IF{$\texttt{dot.rhs} \neq \texttt{dot.lhs}$ and both operands \textit{are}  splittable}
    \STATE Exit and continue traversing the data-flow graph.
\ENDIF
\STATE Let $\texttt{operand\_to\_split} = \texttt{dot.rhs}$ or $\texttt{operand\_to\_split} = \texttt{dot.lhs}$ depending on previous splittability conditions.
\STATE Let $\texttt{split\_dims} = \{d_1, \dots, d_n\}$, $\texttt{split\_producers} = \{\texttt{op}_1, \dots, \texttt{op}_m\}$, s.t.
\STATE\hskip0.4cm $\circ$  $\forall\texttt{op} \in \texttt{split\_producers}$ exists a path $\texttt{op} \rightsquigarrow \texttt{operand\_to\_split}$
\STATE\hskip0.4cm $\circ$  $\forall\texttt{op} \in \texttt{split\_producers}$, $\texttt{input\_size(op)} \le \texttt{tensor\_size\_threshold}$
\STATE\hskip0.4cm $\circ$  $\forall\texttt{op} \in \texttt{split\_producers}$, $\exists d \in \texttt{split\_dims}$ which is splittable on the path $\texttt{op} \rightsquigarrow \texttt{operand\_to\_split}$
\IF{$\texttt{split\_dims} = \varnothing$ or $\texttt{split\_producers} = \varnothing$}
    \STATE Exit and continue traversing the data-flow graph.
\ENDIF
\STATE Let $\texttt{best\_split\_dim}=d \in \texttt{split\_dims}$, and $\texttt{ops} \subseteq \texttt{split\_producers}$, s.t.
\STATE\hskip0.4cm $\circ$ $\min_{d \in \texttt{split\_dims}} \lfloor d \divisionsymbol \texttt{split\_size(operand\_to\_split, tensor\_split\_size)} \rfloor$
\STATE\hskip0.4cm $\circ$ $\forall\texttt{op} \in \texttt{ops}$, the path $\texttt{op} \rightsquigarrow \texttt{operand\_to\_split}$ is splittable on $\texttt{best\_split\_dim}$
\STATE Let $\texttt{split\_size}= \lfloor \texttt{best\_split\_dim} \divisionsymbol \texttt{split\_size(operand\_to\_split, tensor\_split\_size)} \rfloor$

\STATE Create while loop \texttt{HloInstruction}, s.t.
\STATE\hskip0.4cm $\circ$ The loop iterates splits of size \texttt{split\_size} at \texttt{best\_split\_dim} of paths $\texttt{ops} \rightsquigarrow \texttt{operand\_to\_split}$
\STATE\hskip0.4cm $\circ$ The loop applies \texttt{dot} reduction operation on the slice of \texttt{operand\_to\_split}
\STATE\hskip0.4cm $\circ$ The slice result of the \texttt{dot} reduction operation is put into the replica of the original \texttt{dot.result}
\STATE Replace $\texttt{ops} \rightsquigarrow \texttt{dot.result}$ instructions with created while loop.
\ENDPROCEDURE
\end{algorithmic}
\end{algorithm}

Often, a part of a computational data-flow graph can be divided into multiple independent copies, such that each copy of the data-flow graph or its part act on a slice of the input tensor, and the results are combined afterwards in some fashion. This splitting approach is also known as a MapReduce technique \citep{dean2008mapreduce}, where a computation is divided into smaller and less expensive parts (map) and then combined into the final result (reduce). The splitting technique is common for distributing the computational load. The focus of existing solutions is on exploiting hardware parallelism or utilising multiple devices. Instead, we use the same techniques for reducing total memory consumption, which is possible because the memory for individual map operations can be freed before the whole result is computed.

\begin{figure}[h!]
    \centering
    \input{tikz/scheme}
    \caption{The scheme demonstrates transformation of the data-flow graph on the left to the data-flow graph on the right. The graph on the left consists of \texttt{G} and \texttt{R} blocks which are generator and reducer operations respectively, \texttt{I} is the tensor input of \texttt{G}, \texttt{A} is the tensor output of \texttt{G} and \texttt{O} is the tensor output of \texttt{R}. The bracket notation \texttt{A[..., a, ...]} means that the tensor $\texttt{A}$ has a dimension of size $\texttt{a}$ and $\texttt{A}$ can have other dimensions. The \texttt{i:j} is a slicing operation. $\texttt{A} \rightsquigarrow \texttt{R}$ denotes an arbitrary amount of operations in a computational path of the data-flow graph between a tensor \texttt{A} and an operation \texttt{R}. The eXLA splitting optimisation procedure converts the left graph into the loop of independent iterations performing the same chain of operations on a small slice \texttt{i:j}.}
    \label{fig:split-scheme}
\end{figure}
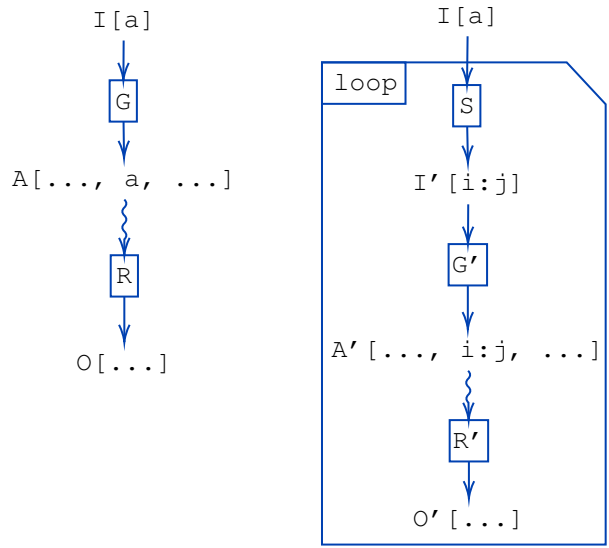

An optimisation pass starts by running a depth-first search from the final result of the computation. The operations \texttt{dot} or \texttt{reduce\_*} are special, as they often indicate that a computation involving a large tensor can give a smaller result. Once a \texttt{dot} or \texttt{reduce\_*} operation is marked as fully traversed, we recursively search the traversed paths for operands that are impractically large tensors, until we reach operands that are deemed small enough. Along the way, we keep track of which operations are applied, and along which axes they are trivially parallelisable. The result is a collection of sub-graphs, that start at operations that produce large tensors, and end at operations that reduce them again, together with candidate axes that they can be split across. According to some heuristics which ensure appropriate sizes for intermediate results, we then turn this entire computation into a while loop, where each iteration computes a manageable part of the final result (\cref{fig:split-scheme}).

Checking if the axis is splittable is necessary as not all operations act independently on each dimension. For example, element-wise operations can be split on any axis, whereas the triangular solve operation can be split on ``batching'' dimensions only. Next, the data-flow graph splitting procedure selects the dimension of the largest size which contributes the most to memory.

As we discussed earlier, the decision about where to split the graph depends on the tensor size. We offer two XLA options to the user for out-of-memory mitigation: \textit{tensor size threshold} and \textit{tensor split size upper bound}. Tensor size threshold is a criterion designed for detecting which operations should be marked as candidates for splitting. Tensor split size upper bound serves as a threshold on the largest allowed chunk size for splitting. These options are set equal by default. The command-line snippet at \cref{code:exla-options} shows how a user would use these options by passing them via an environment variable, and the snippet is indifferent to the machine learning framework used by the script. Minimal user effort is required for using our XLA compiler extension. The user is involved only in defining what the suitable threshold and splitting sizes are.

One strong benefit of our compiler-based solution, is that the computational graph represents the whole pipeline of computations, including forward and backward propagation. Our splitting procedure will be applied automatically, regardless of how many derivatives need to be computed. In addition, our procedure encompasses two splitting schemes that the machine laerning literature distinguishes: model-based and data-based splitting schemes of the data-flow graph. The model-based splitting scheme involves partitioning the model over its parameters, whereas the data-based splitting scheme batches over inputs and, therefore, an algorithm. The proposed splitting approach is suited for supporting both schemes out of the box.

\begin{listing}[b!]
\begin{minted}{bash}
XLA_FLAGS="--xla_tensor_size_threshold=1GB --xla_tensor_split_size=500MB" \
python train.py
\end{minted}
\caption{Example of how a user can set options for the extended XLA using the environment variable.}
\label{code:exla-options}
\end{listing}

\subsection{XLA limitations}
\label{ssec:xla-limitations}
While we still believe that XLA is the right framework for our extensions, several limitations came to light during implementation.

One limitation that is shared with all current frameworks, is that they only have a weak linear algebra type system, where matrices are represented as arrays without additional properties. Solutions that support stronger type systems \citep{bezanson2017julia,barthels2021linnea} may be able to implement a wider variety of match-and-replace optimisations.

Another limitation comes from the default memory allocation manager not being aware of memory limits. Its current behaviour is to execute nodes in the computational graph, and therefore allocate any required memory, as soon as the required inputs have been computed. This means that even if tensors are split to manageable sizes, memory overflows can still occur if several are executed simultaneously. To prevent this from happening, we had to use memory limits that were smaller than our total GPU memory.

\section{Experiments}
\label{sec:experiments}
This section shows how existing software packages take advantage of our extension to XLA (eXLA). We demonstrate our optimisations on non-parametric k-nearest neighbours (kNN) and sparse Gaussian process regression (SGPR) models. 

\begin{figure}[t]
    \centering
    \includegraphics[width=0.95\textwidth]{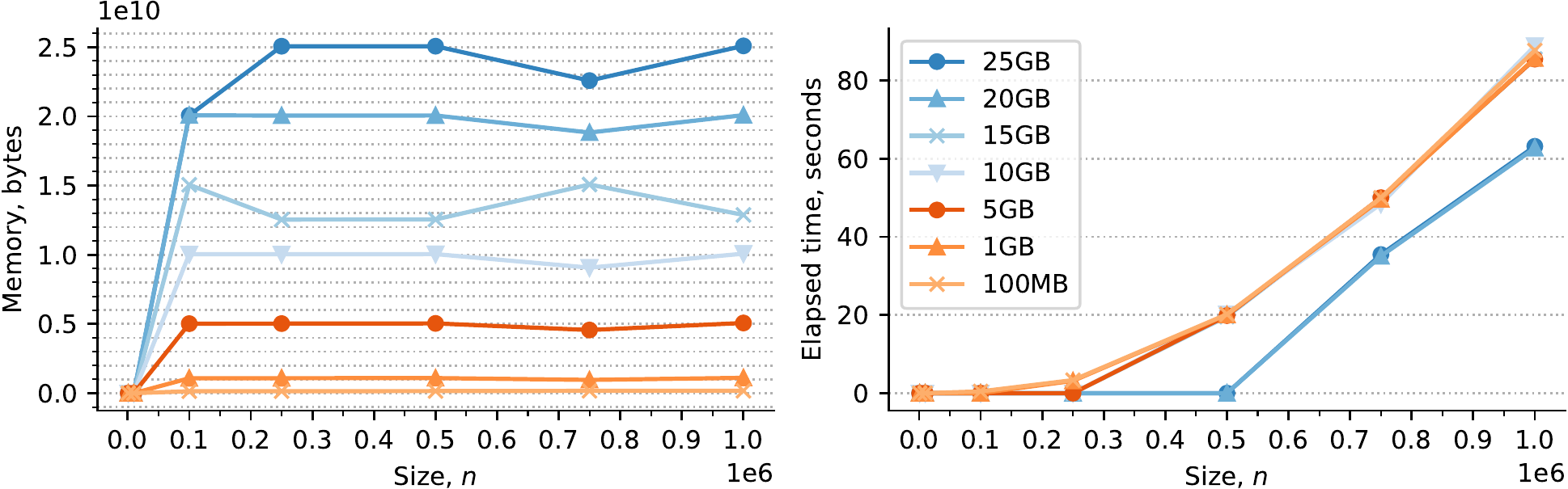}
    \caption{GPU memory consumption and elapsed time of $n\times n$ squared exponential kernel matrix-vector multiplication.}
    \label{fig:mem-consumption}
\end{figure}

\subsection{Matrix-Vector Multiplication}
\label{subsec:mem-control}

We start by demonstrating the improved efficiency that eXLA offers to large-scale matrix-vector multiplications of the form $y = \nKmat v$, where $\nKmat$ is an $n\times n$ kernel matrix, and $y, v \in \R^{n}$. Such computations are common in Conjugate-Gradients-based Gaussian process approximations \citep{gibbs1997efficient,wang2019exact,artemev2021tighter}, where $\nKmat_{ij} = k(x_i, y_j)$ and $k$ is some kernel function. We choose the common Squared Exponential.

We implement this equation using GPflow \citep{gpflow2017}, a TensorFlow-based package that provides a convenient software interface for Gaussian processes and kernel functions. Without eXLA, the entire $\nKmat$ would be stored in memory, leading to a $n^2$ memory cost. This makes running on large datasets infeasible, where e.g.~$n=10^6$ would lead to a memory requirement of 8TB, which is impractical even for the largest of modern GPUs with 40GB of memory. 
A memory efficient split/implicit implementation is necessary to scale to large datasets, as was impressively done by \citet{wang2019exact}, but is cumbersome.

We ran our implementation with eXLA enabled, which allows a user to control the memory of an algorithm. We evaluated the expression in double precision on a Tesla V100 GPU with 32 GB of memory, and applied a range of memory limits. In \cref{fig:mem-consumption} we report the peak memory consumption and execution time of evaluating the kernel matrix-vector product for different sizes, with different memory limits applied. We see that the memory constraints are not violated, and that dataset sizes are used that are far beyond the 32 GB memory capacity.

\subsection{K-Nearest Neighbours}
\label{subsec:knn}
\begin{table*}[t!]
    \centering
    \input{tables/knn-short}
    \caption{Query processing rates (queries per second) for kNN. $n$ and $d$ are the number of data points and the data dimension respectively. Runs which failed due to memory overflow are denoted by $\varnothing$. Runs with eXLA are denoted eJAX and eTF respectively.}
    \label{tab:knn}
\end{table*}

K-nearest neighbours is a fundamental machine learning algorithm, with a similar large memory cost. A kNN query selects $k$ closest data points in the dataset to each query point. Brute-force implementations compute pairwise distances between $m$ query points and $n$ data points, resulting in the distance matrix of size $m \times n$. This is followed by a \texttt{topk} operation, which is often naively implemented using column-wise \texttt{sort} operation on the distance matrix. Our benchmarks show that eXLA scales the brute-force approach and does not fail for large problems, i.e.~large $n$ and $m$.

We compare TensorFlow and JAX implementations with and without eXLA optimisations, and a KeOps implementation. We use randomly generated data, common benchmarks like MNIST and Fashion-MNIST, and Glove-50, Glove-100 and Glove-200 from the ANN-benchmark toolkit \cite{aumuller2020ann}. We use $m=\num{1e4}$ query points in all benchmarks.

Our results are listed in \cref{tab:knn} (see the appendix for a full table that reproduces \citet[table 3]{feydy2020fast}). In all benchmarks, we set the tensor size threshold for eXLA to 100MB for simplicity, even though this may not be optimal for performance. We observe that eXLA prevents memory overflows in JAX and TensorFlow. In addition, performance is comparable or higher. We acknowledge that KeOps performs significantly better than any JAX or TensorFlow implementation. This is explained by \textbf{1)} JAX/TF not having efficient implementations for certain functions (e.g.~\texttt{topk} runs a full sorting algorithm), and \textbf{2)} KeOps having implemented additional optimisations, which could also be added to XLA. However, we note that we also achieved our goal of improving the memory and time performance of a JAX/TensorFlow implementation \textit{without changing the code}.

\begin{figure*}[b!]
    \centering
    \includegraphics[width=0.95\textwidth]{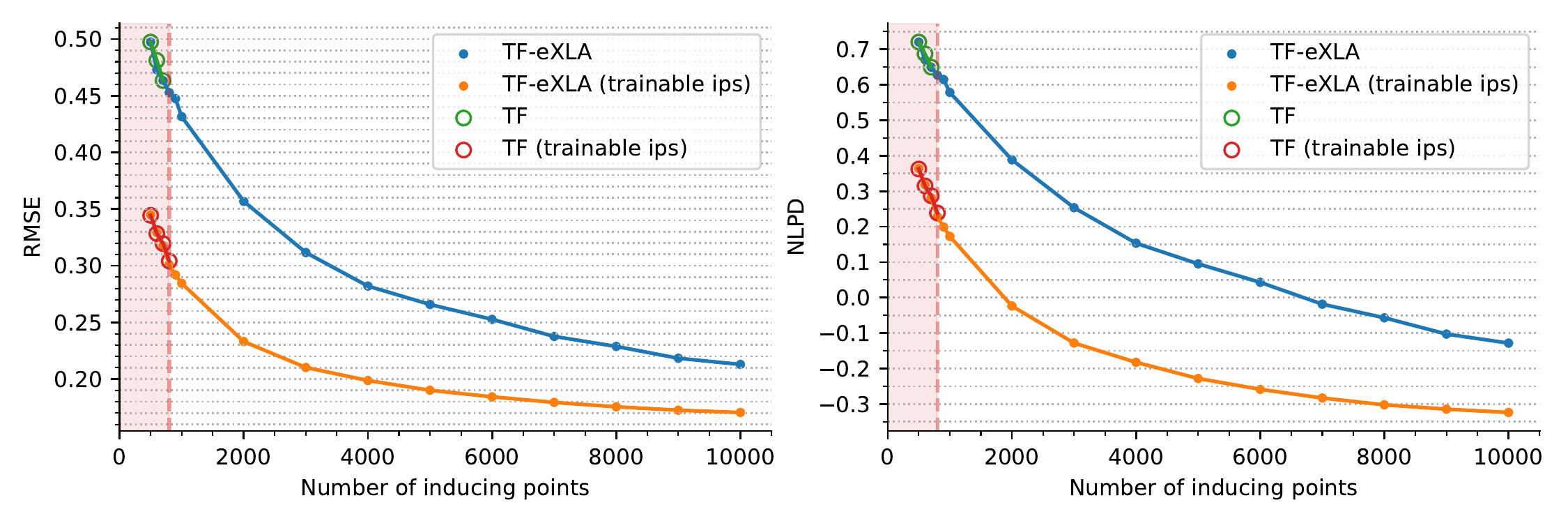}
    \caption{Root mean squared error (RMSE) and negative log predictive density (NLPD) performance test metrics of SGPR for \texttt{3droad} dataset as the number of inducing points is increased. The red shaded region emphasizes the capacity of the SGPR model which user can run using standard GPflow and TensorFlow release packages.
    }
    \label{fig:increasing-ips}
\end{figure*}

\subsection{Sparse Gaussian Process Regression}
\label{subsec:sgpr}
Gaussian processes \citep{rasmussen2006gaussian} are considered the gold standard method for performing regression with uncertainty estimates. A straightforward implementation requires taking a matrix decomposition of an $n\times n$ kernel matrix (like those considered in \cref{subsec:mem-control}), which leads to an $O(n^3)$ time cost, and an $O(n^2)$ memory cost. Scaling Gaussian process is challenging, which is often attributed to the time cost. In reality however, large datasets cause memory overflows far before long runtimes become an obstacle.

Approximate methods have been introduced to deal with both the time and space issues. While there are many variants \citep{quinonero2005unifying}, we consider the sparse variational approximation \citep{titsias2009variational} for which a naive implementation has $O(nm^2 + m^3)$ time cost, and $O(nm + m^2)$ memory cost. Here, $m$ denotes the number of \textit{inducing variables}, which controls the quality of the approximation. Under certain conditions, the method provides reliable hyperparameter selection \citep{bauer2016understanding}, and very accurate posterior approximations \citep{burt2019rates,burt2020convergence} while using $m \ll n$. In practice, these standard implementations may still have their performance limited by how large $m$ can become before a memory overflow occurs. A more memory-efficient implementation with a memory cost of $O(m^2)$ does exist \citep{gal2014distributed}, but is so cumbersome to implement that it is not widely used or compared against.

Fortunately, the splitting optimisation we implemented in eXLA can discover the same procedure that was engineered by \citet{gal2014distributed}. Moreover, since eXLA operates on the entire computation graph, it optimises gradients as well as the optimisation objective function with no additional effort, while \citet{gal2014distributed} needed to implement gradients manually. We demonstrate the utility of eXLA by scaling the GPflow \citep[2.3.1 release version]{gpflow2017} implementation of Sparse Gaussian process regression \citep[SGPR, ][]{titsias2009variational}, \textit{without any modifications} of the code.

With our eXLA optimisations, SGPR was able to scale to much larger datasets, with more inducing points. We conduct experiments on a Tesla V100 GPU with 32 GB of memory, and run on two of the largest UCI datasets that are commonly considered in Gaussian process research: \texttt{3droad} and \texttt{houseelectric}. We primarily compare to \citet{wang2019exact}, who use a Conjugate Gradients approximation \citep{gibbs1997efficient} to achieve the most impressive scaling of a Gaussian process approximation to date, using an impressively engineered implementation that manually splits and distributes parts of the computation.

In \cref{fig:increasing-ips} we compare GPflow's SGPR implementation with and without eXLA as we increase the number of inducing points. We see that until about 800 inducing points, the normal and eXLA runs result in the same predictive metrics, as desired. After 800 inducing points, runs without XLA fail with an ``out of memory'' error, while with eXLA we scaled to $10^4$ inducing points. Simply scaling the method in this way leads to significant performance improvements.

We now compare predictive accuracies directly with the scalable Conjugate Gradients implementation of \citet{wang2019exact}. In that paper, SGPR was discussed as a method that would not scale, probably due to the difficulty of implementing it in a memory-efficient way as in \citet{gal2014distributed}. \Cref{tab:sgpr-houseelectric} shows that using eXLA to scale SGPR can improve predictive performance to such a degree that it can outperform the Conjugate Gradients implementation of \citet{wang2019exact}, without needing additional hardware.

\begin{table*}[b!]
    \centering
    \input{tables/sgpr}
    \caption{SGPR performance on \texttt{houseelectric} and \texttt{3droad} dataset. Iterative GP* and Iterative GP** are trained with lengthscale per dimension and shared lengthscale across dimensions respectively. Iterative GP values are from \cite{wang2019exact}, with unreported metrics denoted as $\varnothing$.}
    \label{tab:sgpr-houseelectric}
\end{table*}

\section{Discussion}
\label{sec:discussion}
We showed that our additional XLA compiler optimisation passes (eXLA) could manage memory overflows algorithms with large tensor or linear algebra operations. The developed compiler extension automatically adjusts computational data-flow graphs to control memory utilisation. As demonstrated in the experiments section, we successfully ran machine learning models compiled with eXLA on a greater scale, whereas their out-of-the-box implementations failed with Out of Memory errors. Crucially, we used existing software packages without modifying any code.

In addition to showing that our compiler extensions work as intended, our experiments also provide directly useful empirical results for Gaussian processes. We managed to run an ``old'' method \citep[SGPR,][]{titsias2009variational}, with unchanged code, to obtain empirical results that outperformed a state-of-the art method \citep{wang2019exact}. This corrects earlier observations in the literature that these methods are inaccurate, and shows that---if the methods can be scaled---they may behave according to theory that shows that they should provide very accurate solutions \citep{burt2020convergence}.

The exciting possiblity of eXLA is that it opens up the possibility to probe behaviour of machine learning models in regimes that were previously infeasible, and on cheap hardware. For example, one could train very wide neural networks, to empirically compare to behaviour predicted by NTK theory \citep{lee2018wide,matthews2018wide,jacot2018ntk,novak2020}. In addition, transformers \citep{vaswani2017transformer} are notoriously memory hungry, and eXLA could help with running them on cheaper hardware, or distributing them across GPUs, without increasing software complexity.

The current implementation of eXLA is still only a demonstration of what compiler optimisations could achieve, and many more optimisations can be added. We believe that increasing the capability of compilers like XLA will greatly increase the efficiency of researchers and practitioners. We hope that community-driven compiler projects will contribute to the community in a similar way to how existing numerical frameworks already do.

\section{Acknowledgements}
Thanks to David R. Burt and Sebastian W. Ober for the feedback on the draft of this paper. We also would like to thank Lev Walkin and Renat Idrisov for discussions about compilers in the beginning of this project.


\bibliography{main}
\bibliographystyle{apalike}

\appendix

\section{Appendix}
\subsection{Code}
The code for benchmarks and experiments is available at \url{https://github.com/awav/gambit}, and the fork of TensorFlow repository with extension to XLA compiler (eXLA) is available at \url{https://github.com/awav/tensorflow}.
\subsection{Additional experiments}
\begin{table*}[ht!]
    \centering
    \input{tables/knn-long}
    \caption{Query processing rates (queries per second) for kNN. $n$ and $d$ are the number of data points and the data dimension respectively. Runs which failed due to memory overflow are denoted by $\varnothing$. Runs with eXLA are denoted eJAX and eTF respectively.}
    \label{tab:knn}
\end{table*}

\begin{figure}
    \centering
    \includegraphics[width=0.7\textwidth]{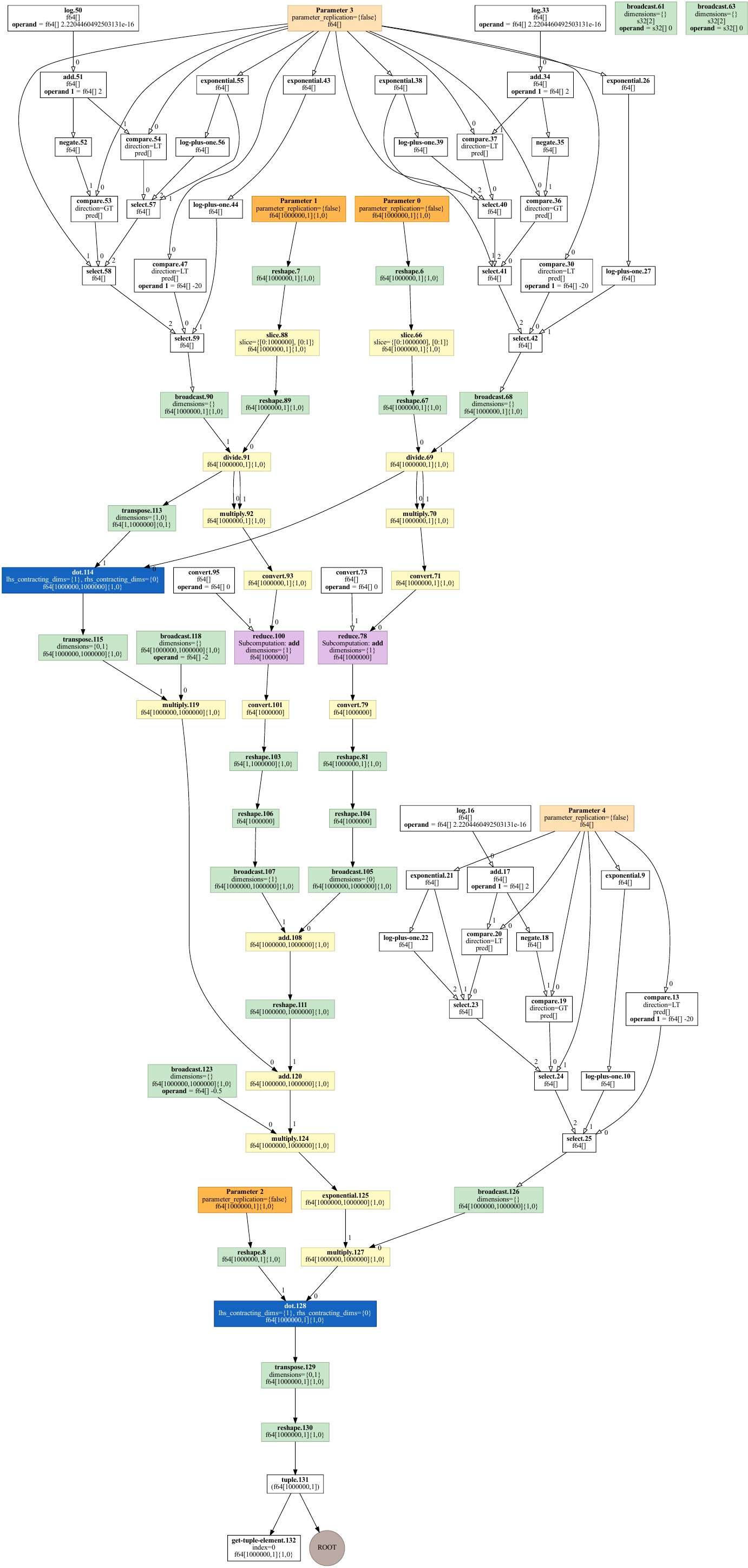}
    \caption{XLA HLO graph for kernel matrix-vector multiplication \textbf{before} splitting optimisation pass is applied in the XLA optimisation pipeline.}
    \label{fig:hlo-before}
\end{figure}

\begin{figure}
    \centering
    \includegraphics[width=1.1\textwidth]{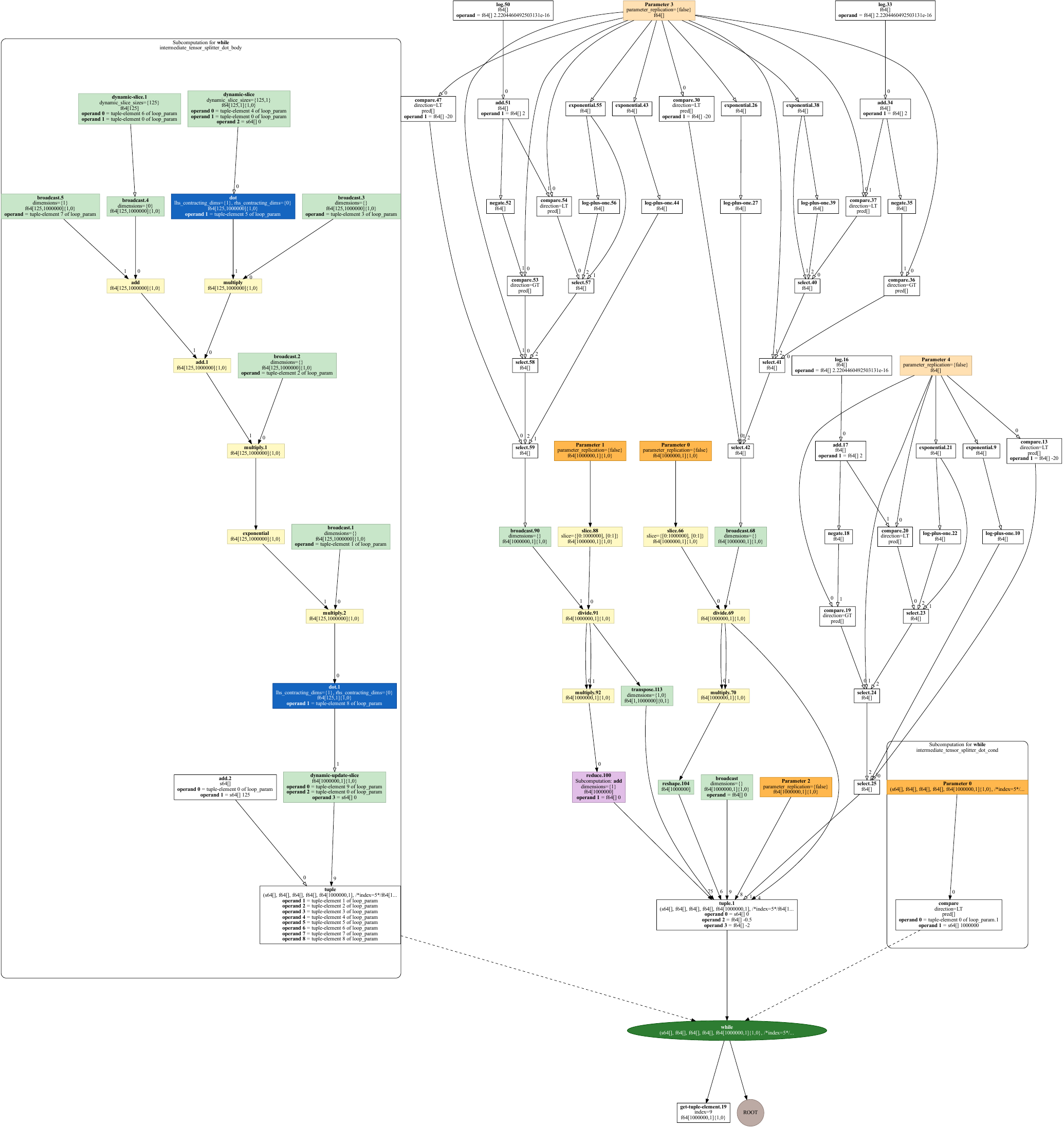}
    \caption{XLA HLO graph for kernel matrix-vector multiplication \textbf{after} splitting optimisation pass is applied in the XLA optimisation pipeline.}
    \label{fig:hlo-after}
\end{figure}

\Cref{fig:hlo-before,fig:hlo-after} depict XLA HLO graphs for kernel matrix-vector multiplication before and after splitting optimisation in eXLA optimisation pipeline (\cref{ssec:splitting}), respectively. The same configuration of the kernel is used as in \cref{subsec:mem-control}, i.e. squared exponential kernel from \cite{gpflow2017}.
By kernel matrix-vector multiplication expression we mean the function $g(\bfx, \bfy, \bfv) = k(\bfx, \bfy) \bfv$, where $k(\bfx, \bfy) = \sigma^2\exp(-1/2 \norm{\bfx - \bfy}^2 / l^2)$ is the kernel with $\sigma^2$ and $l$ hyperparameters.
The size of 1-dimensional input vectors $\bfv$, $\bfx$ and $\bfy$ is $\num{1e-6}$. In turn, the size of the corresponding kernel matrix is $\num{1e-6} \times \num{1e-6}$, and in double precision would require to allocate 8TB. The \textit{tensor size threshold} was set to 1GB, and eXLA splitting optimisation pass divided the expression of the kernel matrix-vector multiplication into smaller chunks, such that the maximum tensor size in the graph is $125 \times \num{1e-6}$.

\end{document}

%% file: tikz/scheme.tex
\begin{tikzpicture}[x=0.75pt,y=0.75pt,yscale=-0.8,xscale=0.8]

\draw  [color={rgb, 255:red, 0; green, 81; blue, 177 }  ,draw opacity=1 ][line width=0.75]  (340,67.33) -- (492.69,67.33) -- (517,93.65) -- (517,370.5) -- (340,370.5) -- cycle ;

\draw  [color={rgb, 255:red, 0; green, 81; blue, 177 }  ,draw opacity=1 ]  (207.5,78.58) -- (224.5,78.58) -- (224.5,104.58) -- (207.5,104.58) -- cycle  ;
\draw (216,91.58) node   [align=left] {{\fontfamily{pcr}\selectfont G}};
\draw (216.67,140.58) node   [align=left] {{\fontfamily{pcr}\selectfont A[..., a, ...]}};
\draw (216.33,40.58) node   [align=left] {{\fontfamily{pcr}\selectfont I[a]}};
\draw  [color={rgb, 255:red, 0; green, 81; blue, 177 }  ,draw opacity=1 ]  (208.17,188.58) -- (225.17,188.58) -- (225.17,214.58) -- (208.17,214.58) -- cycle  ;
\draw (216.67,201.58) node   [align=left] {{\fontfamily{pcr}\selectfont R}};
\draw (216.67,256.25) node   [align=left] {{\fontfamily{pcr}\selectfont O[...]}};
\draw  [color={rgb, 255:red, 0; green, 81; blue, 177 }  ,draw opacity=1 ]  (419,181.58) -- (443,181.58) -- (443,207.58) -- (419,207.58) -- cycle  ;
\draw (431,194.58) node   [align=left] {{\fontfamily{pcr}\selectfont G'}};
\draw (431,248.25) node  [font=\normalsize] [align=left] {{\fontfamily{pcr}\selectfont A'[..., i:j, ...]}};
\draw (431.33,143.58) node   [align=left] {{\fontfamily{pcr}\selectfont I'[i:j]}};
\draw  [color={rgb, 255:red, 0; green, 81; blue, 177 }  ,draw opacity=1 ]  (419.67,292.25) -- (443.67,292.25) -- (443.67,318.25) -- (419.67,318.25) -- cycle  ;
\draw (431.67,305.25) node   [align=left] {{\fontfamily{pcr}\selectfont R'}};
\draw (431.67,354.58) node   [align=left] {{\fontfamily{pcr}\selectfont O'[...]}};
\draw (431,37.92) node   [align=left] {{\fontfamily{pcr}\selectfont I[a]}};
\draw  [color={rgb, 255:red, 0; green, 81; blue, 177 }  ,draw opacity=1 ]  (421.83,81.58) -- (438.83,81.58) -- (438.83,107.58) -- (421.83,107.58) -- cycle  ;
\draw (430.33,94.58) node   [align=left] {{\fontfamily{pcr}\selectfont S}};
\draw  [color={rgb, 255:red, 0; green, 81; blue, 177 }  ,draw opacity=1 ]  (340,67.33) -- (392.12,67.33) -- (392.12,93.4) -- (340,93.4) -- cycle  ;
\draw (389.12,89.4) node [anchor=south east] [inner sep=0.75pt]   [align=left] {{\fontfamily{pcr}\selectfont loop}};
\draw [color={rgb, 255:red, 0; green, 81; blue, 177 }  ,draw opacity=1 ]   (216.25,53.58) -- (216.1,76.58) ;
\draw [shift={(216.08,78.58)}, rotate = 270.37] [color={rgb, 255:red, 0; green, 81; blue, 177 }  ,draw opacity=1 ][line width=0.75]    (10.93,-3.29) .. controls (6.95,-1.4) and (3.31,-0.3) .. (0,0) .. controls (3.31,0.3) and (6.95,1.4) .. (10.93,3.29)   ;
\draw [color={rgb, 255:red, 0; green, 81; blue, 177 }  ,draw opacity=1 ]   (216.18,104.58) -- (216.46,125.58) ;
\draw [shift={(216.49,127.58)}, rotate = 269.22] [color={rgb, 255:red, 0; green, 81; blue, 177 }  ,draw opacity=1 ][line width=0.75]    (10.93,-3.29) .. controls (6.95,-1.4) and (3.31,-0.3) .. (0,0) .. controls (3.31,0.3) and (6.95,1.4) .. (10.93,3.29)   ;
\draw [color={rgb, 255:red, 0; green, 81; blue, 177 }  ,draw opacity=1 ]   (216.67,153.58) .. controls (218.34,155.25) and (218.34,156.91) .. (216.67,158.58) .. controls (215,160.25) and (215,161.91) .. (216.67,163.58) .. controls (218.34,165.25) and (218.34,166.91) .. (216.67,168.58) .. controls (215,170.25) and (215,171.91) .. (216.67,173.58) .. controls (218.34,175.25) and (218.34,176.91) .. (216.67,178.58) -- (216.67,178.58) -- (216.67,186.58) ;
\draw [shift={(216.67,188.58)}, rotate = 270] [color={rgb, 255:red, 0; green, 81; blue, 177 }  ,draw opacity=1 ][line width=0.75]    (10.93,-3.29) .. controls (6.95,-1.4) and (3.31,-0.3) .. (0,0) .. controls (3.31,0.3) and (6.95,1.4) .. (10.93,3.29)   ;
\draw [color={rgb, 255:red, 0; green, 81; blue, 177 }  ,draw opacity=1 ]   (216.67,214.58) -- (216.67,241.25) ;
\draw [shift={(216.67,243.25)}, rotate = 270] [color={rgb, 255:red, 0; green, 81; blue, 177 }  ,draw opacity=1 ][line width=0.75]    (10.93,-3.29) .. controls (6.95,-1.4) and (3.31,-0.3) .. (0,0) .. controls (3.31,0.3) and (6.95,1.4) .. (10.93,3.29)   ;
\draw [color={rgb, 255:red, 0; green, 81; blue, 177 }  ,draw opacity=1 ]   (431.67,318.25) -- (431.67,339.58) ;
\draw [shift={(431.67,341.58)}, rotate = 270] [color={rgb, 255:red, 0; green, 81; blue, 177 }  ,draw opacity=1 ][line width=0.75]    (10.93,-3.29) .. controls (6.95,-1.4) and (3.31,-0.3) .. (0,0) .. controls (3.31,0.3) and (6.95,1.4) .. (10.93,3.29)   ;
\draw [color={rgb, 255:red, 0; green, 81; blue, 177 }  ,draw opacity=1 ]   (431.15,261.25) .. controls (432.84,262.9) and (432.86,264.56) .. (431.21,266.25) .. controls (429.56,267.94) and (429.58,269.6) .. (431.27,271.25) .. controls (432.96,272.9) and (432.98,274.56) .. (431.33,276.25) .. controls (429.68,277.94) and (429.7,279.6) .. (431.39,281.25) -- (431.4,282.25) -- (431.49,290.25) ;
\draw [shift={(431.51,292.25)}, rotate = 269.33] [color={rgb, 255:red, 0; green, 81; blue, 177 }  ,draw opacity=1 ][line width=0.75]    (10.93,-3.29) .. controls (6.95,-1.4) and (3.31,-0.3) .. (0,0) .. controls (3.31,0.3) and (6.95,1.4) .. (10.93,3.29)   ;
\draw [color={rgb, 255:red, 0; green, 81; blue, 177 }  ,draw opacity=1 ]   (431.25,156.58) -- (431.1,179.58) ;
\draw [shift={(431.08,181.58)}, rotate = 270.37] [color={rgb, 255:red, 0; green, 81; blue, 177 }  ,draw opacity=1 ][line width=0.75]    (10.93,-3.29) .. controls (6.95,-1.4) and (3.31,-0.3) .. (0,0) .. controls (3.31,0.3) and (6.95,1.4) .. (10.93,3.29)   ;
\draw [color={rgb, 255:red, 0; green, 81; blue, 177 }  ,draw opacity=1 ]   (431,207.58) -- (431,233.25) ;
\draw [shift={(431,235.25)}, rotate = 270] [color={rgb, 255:red, 0; green, 81; blue, 177 }  ,draw opacity=1 ][line width=0.75]    (10.93,-3.29) .. controls (6.95,-1.4) and (3.31,-0.3) .. (0,0) .. controls (3.31,0.3) and (6.95,1.4) .. (10.93,3.29)   ;
\draw [color={rgb, 255:red, 0; green, 81; blue, 177 }  ,draw opacity=1 ]   (430.85,50.92) -- (430.51,79.58) ;
\draw [shift={(430.49,81.58)}, rotate = 270.67] [color={rgb, 255:red, 0; green, 81; blue, 177 }  ,draw opacity=1 ][line width=0.75]    (10.93,-3.29) .. controls (6.95,-1.4) and (3.31,-0.3) .. (0,0) .. controls (3.31,0.3) and (6.95,1.4) .. (10.93,3.29)   ;
\draw [color={rgb, 255:red, 0; green, 81; blue, 177 }  ,draw opacity=1 ]   (430.6,107.58) -- (431.03,128.58) ;
\draw [shift={(431.07,130.58)}, rotate = 268.83] [color={rgb, 255:red, 0; green, 81; blue, 177 }  ,draw opacity=1 ][line width=0.75]    (10.93,-3.29) .. controls (6.95,-1.4) and (3.31,-0.3) .. (0,0) .. controls (3.31,0.3) and (6.95,1.4) .. (10.93,3.29)   ;

\end{tikzpicture}

%% file: tables/knn-short.tex
\begin{tabular}{lllrrrrrrr}
\toprule
   Dataset & Distance &         n &    d &   KeOps & eJAX & eTF  &          JAX &            TF \\
\midrule
    Random &       $L^2$ &    1e4 &  100 &  983263 &   277364 & 284777 &        281695 &        280826 \\ 
    Random &       $L^2$ &    1e4 &    3 & 3662188 &   292804 & 294971 &        288098 &        294776 \\ 
    Random &       $L^2$ &    1e6 &  100 &   24367 &     2433 &   2530 & $\varnothing$ & $\varnothing$ \\ 
    Random &       $L^2$ &    1e6 &    3 &  123765 &     2512 &   2605 & $\varnothing$ & $\varnothing$ \\ 
\midrule
     MNIST &       $L^2$ & 6e4 &  784 &   41084 &    32290 &  33455 &         25544 &         26138 \\ 
     MNIST &       $L^1$ & 6e4 &  784 &   40697 &     2356 &   2985 &          2498 &          2988 \\ 
\midrule
   Fashion &       $L^2$ & 6e4 &  784 &   40399 &    32382 &  33428 &         25558 &         26128 \\ 
   Fashion &       $L^1$ & 6e4 &  784 &   40982 &     2357 &   2984 &          2498 &          2989 \\ 
\midrule
  Glove-50 &   Cosine &   1.18e6 &   50 & 3464257 &     2103 &   1929 & $\varnothing$ & $\varnothing$ \\
 Glove-100 &   Cosine &   1.18e6 &  100 &  631420 &     2053 &   1871 & $\varnothing$ & $\varnothing$ \\ 
 Glove-200 &   Cosine &   1.18e6 &  200 &  398293 &     1967 &   1724 & $\varnothing$ & $\varnothing$ \\ 
\bottomrule
\end{tabular}

%% file: tables/sgpr.tex
\begin{tabular}{llcccc}
\toprule
   Dataset & Model & RMSE & NLPD & Time (hours) & GPUs \\ 
\midrule
       & SGPR-1000 & $0.048\pm\num{2e-4}$ & $-1.602\pm\num{3e-3}$ & $5.01\pm0.06$ & 1 \\ 
       & SGPR-2000 & $0.046\pm\num{1e-4}$ & $-1.651\pm\num{3e-3}$ & $18.03\pm0.09$ & 1 \\ 
       \texttt{houseelectric} & SGPR-3000 & $0.044\pm\num{1e-4}$ & $-1.696\pm\num{5e-3}$ & $38.68\pm0.14$ & 1 \\ 
       & SGPR-4000 & $\mathbf{0.043}\pm\num{1e-4}$ & $\mathbf{-1.717}\pm\num{5e-3}$ & $50.00\pm0.10$ & 1 \\ 
       & Iterative GP* & $0.054\pm0.000$ & $-0.207\pm0.001$ & $1.55\pm0.02$ & 8 \\
       & Iterative GP** & $0.050$ & $\varnothing$ & $79.96$ & 8 \\ 
\midrule
       & SGPR-1000 & $0.285\pm0.002$ & $-0.173\pm0.004$ & $1.11\pm0.01$ & 1 \\ 
       & SGPR-5000 & $0.190\pm0.002$ & $-0.228\pm0.002$ & $11.33\pm0.03$ & 1 \\ 
       \texttt{3droad} & SGPR-8000 & $0.176\pm0.001$ & $-0.302\pm0.004$ & $28.21\pm0.05$ & 1 \\ 
   & SGPR-10000 & $0.170\pm0.001$ & $\mathbf{-0.322}\pm0.002$ & $41.83\pm0.03$ & 1 \\ 
       & Iterative GP* & $0.110\pm0.017$ & $1.239\pm0.025$ & $1.00\pm\num{2e-3}$ & 8 \\ 
       & Iterative GP** & $\mathbf{0.106}$ & $\varnothing$ & $7.06$ & 8 \\ 
\bottomrule
\end{tabular}

%% file: tables/knn-long.tex
\begin{tabular}{lllrrrrrrr}
\toprule
   Dataset & Distance &         n &    d &   KeOps & eJAX  & eTF &          JAX &            TF \\
\midrule
    Random &       $L^2$ &     10000 &  100 &  983263 &   277364 & 284777 &        281695 &        280826 \\ 
    Random &       $L^2$ &     10000 &   10 & 2587001 &   291751 & 295029 &        287958 &        294168 \\ 
    Random &       $L^2$ &     10000 &    3 & 3662188 &   292804 & 294971 &        288098 &        294776 \\ 
    Random &       $L^2$ &   1000000 &  100 &   24367 &     2433 &   2530 & $\varnothing$ & $\varnothing$ \\ 
    Random &       $L^2$ &   1000000 &   10 &  106726 &     2505 &   2601 & $\varnothing$ & $\varnothing$ \\ 
    Random &       $L^2$ &   1000000 &    3 &  123765 &     2512 &   2605 & $\varnothing$ & $\varnothing$ \\ 
    Random &       $L^2$ &  10000000 &  100 &    2461 &      243 &    253 & $\varnothing$ & $\varnothing$ \\ 
    Random &       $L^2$ &  10000000 &   10 &   11546 &      251 &    261 & $\varnothing$ & $\varnothing$ \\ 
    Random &       $L^2$ &  10000000 &    3 &   13192 &      251 &    261 & $\varnothing$ & $\varnothing$ \\ 
\midrule
    Random &       $L^1$ &   1000000 &  100 &   24307 &      517 &    521 & $\varnothing$ & $\varnothing$ \\ 
    Random &       $L^1$ &   1000000 &   10 &  108739 &     2494 &   2590 & $\varnothing$ & $\varnothing$ \\ 
\midrule
    Random &   Cosine &   1000000 &  100 &   32520 &     2434 &   2515 & $\varnothing$ & $\varnothing$ \\ 
    Random &   Cosine &   1000000 &   10 &  106876 &     2507 &   2612 & $\varnothing$ & $\varnothing$ \\ 
\midrule
     MNIST &       $L^2$ &     60000 &  784 &   41084 &    32290 &  33455 &         25544 &         26138 \\ 
     MNIST &       $L^1$ &     60000 &  784 &   40697 &     2356 &   2985 &          2498 &          2988 \\ 
\midrule
   Fashion &       $L^2$ &     60000 &  784 &   40399 &    32382 &  33428 &         25558 &         26128 \\ 
   Fashion &       $L^1$ &     60000 &  784 &   40982 &     2357 &   2984 &          2498 &          2989 \\ 
\midrule
  Glove-50 &   Cosine &   1183514 &   50 & 3464257 &     2103 &   1929 & $\varnothing$ & $\varnothing$ \\
 Glove-100 &   Cosine &   1183514 &  100 &  631420 &     2053 &   1871 & $\varnothing$ & $\varnothing$ \\ 
 Glove-200 &   Cosine &   1183514 &  200 &  398293 &     1967 &   1724 & $\varnothing$ & $\varnothing$ \\ 
\bottomrule
\end{tabular}

%% file: main.bbl
\begin{thebibliography}{}

\bibitem[Abadi et~al., 2016]{abadi2016tensorflow}
Abadi, M., Barham, P., Chen, J., Chen, Z., Davis, A., Dean, J., Devin, M.,
  Ghemawat, S., Irving, G., Isard, M., et~al. (2016).
\newblock Tensorflow: A system for large-scale machine learning.
\newblock In {\em 12th {USENIX} symposium on operating systems design and
  implementation ({OSDI} 16)}.

\bibitem[Aho et~al., 2006]{aho2006compilers}
Aho, A.~V., Lam, M.~S., Sethi, R., and Ullman, J.~D. (2006).
\newblock {\em Compilers: Principles, Techniques, and Tools (2nd Edition)}.
\newblock Addison-Wesley Longman Publishing Co., Inc.

\bibitem[Artemev et~al., 2021]{artemev2021tighter}
Artemev, A., Burt, D.~R., and Van Der~Wilk, M. (2021).
\newblock Tighter bounds on the log marginal likelihood of {G}aussian process
  regression using conjugate gradients.
\newblock In {\em Proceedings of the 38th International Conference on Machine
  Learning (ICML)}, volume 139.

\bibitem[Aum{\"u}ller et~al., 2020]{aumuller2020ann}
Aum{\"u}ller, M., Bernhardsson, E., and Faithfull, A. (2020).
\newblock {ANN}-{B}enchmarks: A benchmarking tool for approximate nearest
  neighbor algorithms.
\newblock {\em Information Systems}, 87.

\bibitem[Barthels et~al., 2018]{barthels2018generalized}
Barthels, H., Copik, M., and Bientinesi, P. (2018).
\newblock The generalized matrix chain algorithm.
\newblock In {\em Proceedings of the 2018 International Symposium on Code
  Generation and Optimization}.

\bibitem[Barthels et~al., 2021]{barthels2021linnea}
Barthels, H., Psarras, C., and Bientinesi, P. (2021).
\newblock Linnea: Automatic generation of efficient linear algebra programs.
\newblock {\em ACM Transactions on Mathematical Software (TOMS)}, 47.

\bibitem[Bauer et~al., 2016]{bauer2016understanding}
Bauer, M., van~der Wilk, M., and Rasmussen, C.~E. (2016).
\newblock Understanding probabilistic sparse gaussian process approximations.
\newblock In {\em Advances in Neural Information Processing Systems},
  volume~29.

\bibitem[Baydin et~al., 2018]{baydin2018automatic}
Baydin, A.~G., Pearlmutter, B.~A., Radul, A.~A., and Siskind, J.~M. (2018).
\newblock Automatic differentiation in machine learning: a survey.
\newblock {\em Journal of Marchine Learning Research}, 18.

\bibitem[Bezanson et~al., 2017]{bezanson2017julia}
Bezanson, J., Edelman, A., Karpinski, S., and Shah, V.~B. (2017).
\newblock Julia: A fresh approach to numerical computing.
\newblock {\em SIAM review}, 59.

\bibitem[Bradbury et~al., 2018]{jax2018github}
Bradbury, J., Frostig, R., Hawkins, P., Johnson, M.~J., Leary, C., Maclaurin,
  D., Necula, G., Paszke, A., Vander{P}las, J., Wanderman-{M}ilne, S., and
  Zhang, Q. (2018).
\newblock {JAX}: composable transformations of {P}ython+{N}um{P}y programs.

\bibitem[Bronstein et~al., 2017]{bronstein2017geometric}
Bronstein, M.~M., Bruna, J., LeCun, Y., Szlam, A., and Vandergheynst, P.
  (2017).
\newblock Geometric deep learning: going beyond euclidean data.
\newblock {\em IEEE Signal Processing Magazine}, 34.

\bibitem[Burt et~al., 2019]{burt2019rates}
Burt, D., Rasmussen, C.~E., and van~der Wilk, M. (2019).
\newblock {Rates of Convergence for Sparse Variational Gaussian Process
  Regression}.
\newblock In {\em Proceedings of the 36th International Conference on Machine
  Learning (ICML)}, volume~97.

\bibitem[Burt et~al., 2020]{burt2020convergence}
Burt, D.~R., Rasmussen, C.~E., and van~der Wilk, M. (2020).
\newblock Convergence of sparse variational inference in {G}aussian processes
  regression.
\newblock {\em Journal of Machine Learning Research}, 21.

\bibitem[Buyya, 1999]{buyya1999high}
Buyya, R. (1999).
\newblock High performance cluster computing.
\newblock {\em New Jersey: F’rentice}.

\bibitem[Charlier et~al., 2021]{charlier2021keops}
Charlier, B., Feydy, J., Glaunès, J.~A., Collin, F.-D., and Durif, G. (2021).
\newblock Kernel operations on the {GPU}, with autodiff, without memory
  overflows.
\newblock {\em Journal of Machine Learning Research}, 22.

\bibitem[Chen et~al., 2018]{chen2018tvm}
Chen, T., Moreau, T., Jiang, Z., Zheng, L., Yan, E., Shen, H., Cowan, M., Wang,
  L., Hu, Y., Ceze, L., et~al. (2018).
\newblock {TVM}: An automated end-to-end optimizing compiler for deep learning.
\newblock In {\em 13th {USENIX} Symposium on Operating Systems Design and
  Implementation ({OSDI} 18)}.

\bibitem[Chin, 1978]{chin1978n}
Chin, F.~Y. (1978).
\newblock An {O(n)} algorithm for determining a near-optimal computation order
  of matrix chain products.
\newblock {\em Communications of the ACM}, 21.

\bibitem[Czumaj, 1996]{czumaj1996very}
Czumaj, A. (1996).
\newblock Very fast approximation of the matrix chain product problem.
\newblock {\em Journal of Algorithms}, 21.

\bibitem[Dean and Ghemawat, 2008]{dean2008mapreduce}
Dean, J. and Ghemawat, S. (2008).
\newblock Mapreduce: simplified data processing on large clusters.
\newblock {\em Communications of the ACM}, 51.

\bibitem[Feydy et~al., 2020]{feydy2020fast}
Feydy, J., Glaun{\`e}s, J., Charlier, B., and Bronstein, M. (2020).
\newblock Fast geometric learning with symbolic matrices.
\newblock {\em Advances in Neural Information Processing Systems}, 33.

\bibitem[Gal et~al., 2014]{gal2014distributed}
Gal, Y., Van Der~Wilk, M., and Rasmussen, C.~E. (2014).
\newblock Distributed variational inference in sparse {G}aussian process
  regression and latent variable models.
\newblock In {\em Advances in Neural Information Processing Systems},
  volume~27.

\bibitem[Gibbs and Mackay, 1997]{gibbs1997efficient}
Gibbs, M. and Mackay, D. (1997).
\newblock Efficient implementation of {G}aussian processes.
\newblock Technical report, Cavendish Laboratory, University of Cambridge.

\bibitem[Jacot et~al., 2018]{jacot2018ntk}
Jacot, A., Gabriel, F., and Hongler, C. (2018).
\newblock {Neural Tangent Kernel: Convergence and Generalization in Neural
  Networks}.
\newblock In {\em Advances in Neural Information Processing Systems},
  volume~31.

\bibitem[Leary and Wang, 2017]{leary2017xla}
Leary, C. and Wang, T. (2017).
\newblock {XLA}: Tensorflow, compiled.
\newblock {\em TensorFlow Dev Summit}.

\bibitem[Lee et~al., 2018]{lee2018wide}
Lee, J., Sohl-dickstein, J., Pennington, J., Novak, R., Schoenholz, S., and
  Bahri, Y. (2018).
\newblock Deep neural networks as {G}aussian processes.
\newblock In {\em International Conference on Learning Representations}.

\bibitem[Matthews et~al., 2018]{matthews2018wide}
Matthews, A. G. d.~G., Hron, J., Rowland, M., Turner, R.~E., and Ghahramani, Z.
  (2018).
\newblock {G}aussian process behaviour in wide deep neural networks.
\newblock In {\em International Conference on Learning Representations}.

\bibitem[Matthews et~al., 2017]{gpflow2017}
Matthews, A. G. d.~G., {van der Wilk}, M., Nickson, T., Fujii, K.,
  {Boukouvalas}, A., {Le{\'o}n-Villagr{\'a}}, P., Ghahramani, Z., and Hensman,
  J. (2017).
\newblock {{GP}flow: A {G}aussian process library using {T}ensor{F}low}.
\newblock {\em Journal of Machine Learning Research}, 18.

\bibitem[Meanti et~al., 2020]{meanti2020falcon}
Meanti, G., Carratino, L., Rosasco, L., and Rudi, A. (2020).
\newblock Kernel methods through the roof: Handling billions of points
  efficiently.
\newblock In {\em Advances in Neural Information Processing Systems},
  volume~33.

\bibitem[Novak et~al., 2020]{novak2020}
Novak, R., Xiao, L., Hron, J., Lee, J., Alemi, A.~A., Sohl-Dickstein, J., and
  Schoenholz, S.~S. (2020).
\newblock Neural tangents: Fast and easy infinite neural networks in python.
\newblock In {\em International Conference on Learning Representations}.

\bibitem[Paszke et~al., 2019]{paszke2019pytorch}
Paszke, A., Gross, S., Massa, F., Lerer, A., Bradbury, J., Chanan, G., Killeen,
  T., Lin, Z., Gimelshein, N., Antiga, L., et~al. (2019).
\newblock Pytorch: An imperative style, high-performance deep learning library.
\newblock In {\em Advances in Neural Information Processing Systems},
  volume~32.

\bibitem[Qui{{\~n}}onero-Candela and Rasmussen, 2005]{quinonero2005unifying}
Qui{{\~n}}onero-Candela, J. and Rasmussen, C.~E. (2005).
\newblock A unifying view of sparse approximate gaussian process regression.
\newblock {\em Journal of Machine Learning Research}, 6.

\bibitem[Rasmussen and Williams, 2006]{rasmussen2006gaussian}
Rasmussen, C.~E. and Williams, C.~K. (2006).
\newblock {G}aussian processes for machine learning.
\newblock {\em {G}aussian Processes for Machine Learning}.

\bibitem[Rotem et~al., 2018]{rotem2018glow}
Rotem, N., Fix, J., Abdulrasool, S., Catron, G., Deng, S., Dzhabarov, R.,
  Gibson, N., Hegeman, J., Lele, M., Levenstein, R., et~al. (2018).
\newblock Glow: Graph lowering compiler techniques for neural networks.
\newblock {\em arXiv preprint arXiv:1805.00907}.

\bibitem[Schwartz and Weiss, 2019]{schwartz2019revisit}
Schwartz, O. and Weiss, E. (2019).
\newblock Revisiting {``Computation of Matrix Chain Products''}.
\newblock {\em SIAM Journal on Computing}, 48.

\bibitem[Shazeer et~al., 2018]{shazeer2018mesh}
Shazeer, N., Cheng, Y., Parmar, N., Tran, D., Vaswani, A., Koanantakool, P.,
  Hawkins, P., Lee, H., Hong, M., Young, C., et~al. (2018).
\newblock Mesh-tensorflow: Deep learning for supercomputers.
\newblock In {\em Advances in Neural Information Processing Systems},
  volume~31.

\bibitem[Sutton, 2019]{sutton2019bitter}
Sutton, R. (2019).
\newblock The bitter lesson.
\newblock {\em Incomplete Ideas (blog)}, 13.

\bibitem[Titsias, 2009]{titsias2009variational}
Titsias, M. (2009).
\newblock Variational learning of inducing variables in sparse {G}aussian
  processes.
\newblock In {\em Proceedings of the 12th International Conference on
  Artificial Intelligence and Statistics}, volume~5.

\bibitem[Vaswani et~al., 2017]{vaswani2017transformer}
Vaswani, A., Shazeer, N., Parmar, N., Uszkoreit, J., Jones, L., Gomez, A.~N.,
  Kaiser, L.~u., and Polosukhin, I. (2017).
\newblock Attention is all you need.
\newblock In {\em Advances in Neural Information Processing Systems},
  volume~30.

\bibitem[Wang et~al., 2019a]{wang2019exact}
Wang, K., Pleiss, G., Gardner, J., Tyree, S., Weinberger, K.~Q., and Wilson,
  A.~G. (2019a).
\newblock Exact {G}aussian processes on a million data points.
\newblock In {\em Advances in Neural Information Processing Systems},
  volume~32.

\bibitem[Wang et~al., 2019b]{wang2019tofu}
Wang, M., Huang, C.-c., and Li, J. (2019b).
\newblock Supporting very large models using automatic dataflow graph
  partitioning.
\newblock In {\em Proceedings of the 14th EuroSys Conference 2019}.

\end{thebibliography}
